\documentclass{article}
\usepackage{soul}

\usepackage{microtype}
\usepackage{graphicx}
\usepackage{subfigure}
 \usepackage{booktabs} % for professional tables
\usepackage{arydshln}

\usepackage{xcolor}

\definecolor{cc}{rgb}{0.1, 0.2, 0.8}  % 你可以根据需要更改 RGB 值

\newcommand{\cc}[1]{\textcolor{cc}{\textbf{#1}}}
\newcommand{\red}[1]{\textcolor{red}{\textbf{#1}}}

\definecolor{bb}{rgb}{0.3, 0.4, 0.85}  % 更浅的淡蓝色，您可以调整 RGB 值

\newcommand{\takeaway}[1]{\textcolor{bb}{\textit{#1}}}

\usepackage[accepted]{icml2025}
\usepackage{hyperref}
\usepackage{amsmath}
\usepackage{amssymb}
\usepackage{mathtools}
\usepackage{amsthm}
\usepackage{tcolorbox}
\usepackage[capitalize,noabbrev]{cleveref}

\theoremstyle{plain}

\theoremstyle{definition}

\theoremstyle{remark}

\newcommand{\model}{long-output LLMs }

\usepackage[textsize=tiny]{todonotes}

\usepackage{pifont}
\icmltitlerunning{}

\begin{document}

\twocolumn[
% \icmltitle{Position: From Long-context to Long-content: Don't Ignore Long Generation}

\icmltitle{Shifting Long-Context LLMs Research from Input to Output }


\icmlsetsymbol{equal}{*}

\begin{icmlauthorlist}
\icmlauthor{Yuhao Wu}{yyy}
\icmlauthor{Yushi Bai}{thu}
\icmlauthor{Zhiqing Hu}{yyy}
\icmlauthor{Shangqing Tu}{thu}
\icmlauthor{Ming Shan Hee}{yyy}
\icmlauthor{Juanzi Li}{thu}
\icmlauthor{Roy Ka-Wei Lee}{yyy}
\end{icmlauthorlist}

\icmlaffiliation{yyy}{Singapore University of Technology and Design}
\icmlaffiliation{thu}{Tsinghua University}

\icmlcorrespondingauthor{Roy Ka-Wei Lee}{roy\_lee@sutd.edu.sg}
% \newcommand{\model}{long-output LLMs}
\icmlkeywords{Machine Learning, ICML}

\vskip 0.3in]

\printAffiliationsAndNotice{}
% leave blank if no need to mention equal contribution

\begin{abstract}
Recent advancements in long-context Large Language Models (LLMs) have primarily concentrated on processing extended input contexts, resulting in significant strides in long-context comprehension. However, the equally critical aspect of generating long-form outputs has received comparatively less attention. This paper advocates for a paradigm shift in NLP research towards addressing the challenges of long-output generation. Tasks such as novel writing, long-term planning, and complex reasoning require models to understand extensive contexts and produce coherent, contextually rich, and logically consistent extended text. These demands highlight a critical gap in current LLM capabilities. We underscore the importance of this under-explored domain and call for focused efforts to develop foundational LLMs tailored for generating high-quality, long-form outputs, which hold immense potential for real-world applications.
%Recent research on long-context Large Language Models (LLMs) has predominantly focused on the input processing, with considerable attention given to handling extended contexts. However, there has been comparatively less emphasis on the output side, particularly the generation of long-form content. In this position paper, we argue for a shift in research focus within the NLP community towards the critical need for long-output generation. Tasks that involve long generations, such as novel writing, long-term complex planning, and complex reasoning, require models capable of producing extended, coherent, and contextually rich text. These tasks go beyond simply processing lengthy input contexts; they necessitate the generation of extended outputs that are both creative and logically consistent. This paper highlights this significant gap and urges further research into developing \model capable of generating extended outputs.
\end{abstract}
\section{Introduction}

\paragraph{Advancements in Long-Context LLMs (Inputs).}
Research on long-context Large Language Models (LLMs) has progressed rapidly in recent years, particularly in expanding context window lengths. These have grown from an initial 8K tokens to as much as 128K or even 1M tokens~\cite{GPT-4o,claude-3-5,reid2024gemini,glm2024chatglm,dubey2024llama}. This dramatic expansion has enabled significant improvements in performance across long-context benchmarks~\cite{needleinhaystack,bai2024longbench,hsieh2024ruler}, unlocking new possibilities for real-world applications. Such advancements facilitate enhanced long-document and multi-document retrieval and more nuanced text comprehension tasks. For example, applications like summarizing lengthy reports, answering questions based on entire books, and analyzing multi-chapter documents are now increasingly viable~\cite{bai2024longbench,an2024leval,hsieh2024ruler,vodrahalli2024michelangelo,reid2024gemini}. Consequently, the ability to process extended text has evolved from a specialized feature into a fundamental capability of state-of-the-art LLMs.

%\paragraph{Let us start with Long-context LLMs (inputs).} Research and development in long-context Large Language Models have progressed considerably over the past year, with context window lengths expanding significantly. Initially at 8K tokens, context windows have grown to 128K and even 1M tokens~\cite{GPT-4o,claude-3-5,reid2024gemini,glm2024chatglm,dubey2024llama}. This expansion has been accompanied by promising performance on long-context benchmarks~\cite{needleinhaystack,bai2024longbench,hsieh2024ruler}. Long-context LLMs offer substantial potential for diverse real-world applications, including enhanced long-document and multi-document retrieval, as well as more sophisticated text comprehension tasks. These models enable capabilities previously unattainable with traditional LLMs. Furthermore, the ability to process extended text is crucial for tasks such as summarizing lengthy reports, answering questions based on entire books, and analyzing multi-chapter documents~\cite{bai2024longbench,an2024leval,hsieh2024ruler,vodrahalli2024michelangelo,reid2024gemini}. Consequently, the long-context ability has rapidly become an integral component of state-of-the-art LLMs, transitioning from a specialized function to an essential feature of modern language models.

\begin{figure}[t]
    \centering
    \includegraphics[width=0.9\columnwidth]{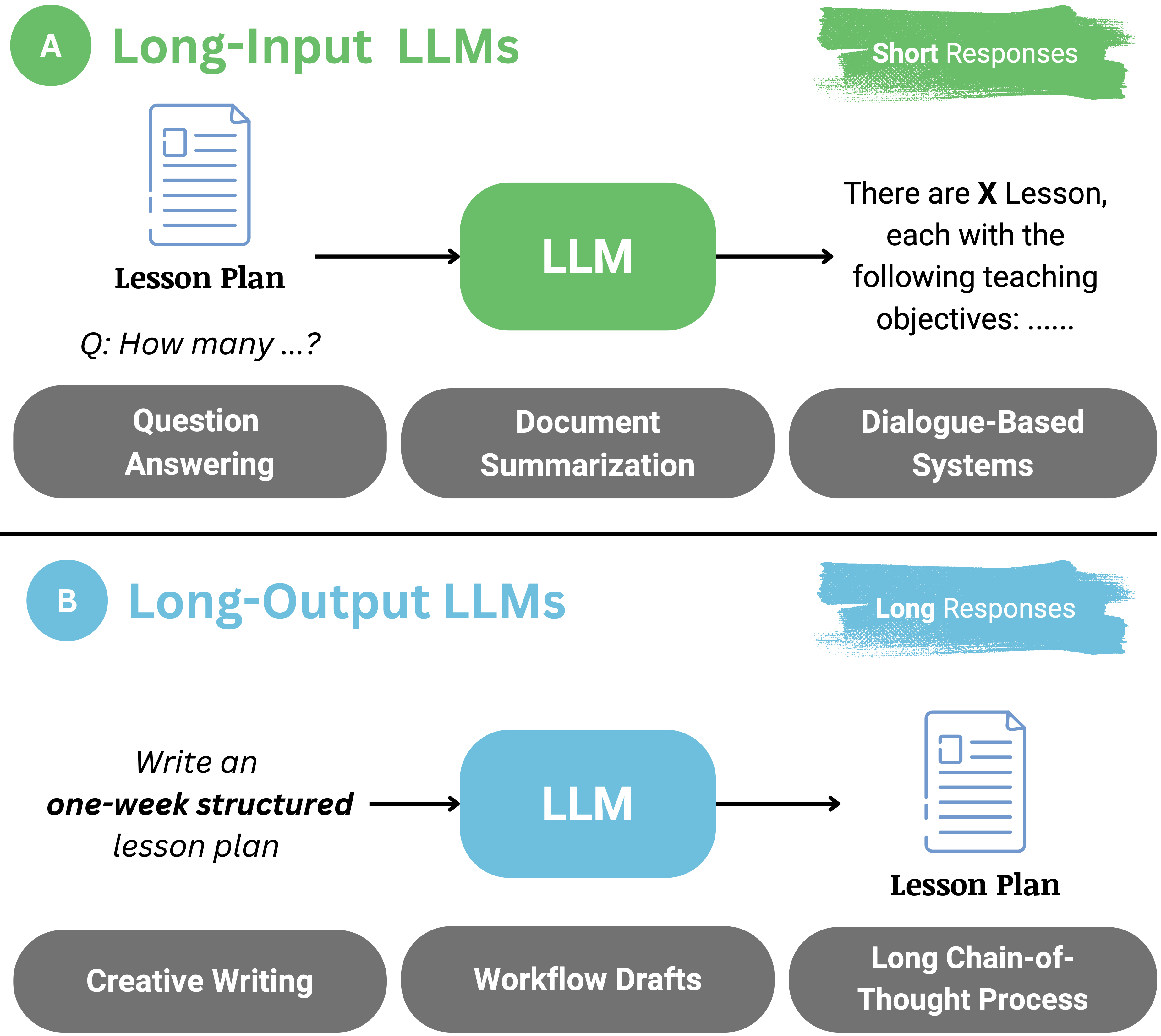}  % LaTeX默认的占位符，适应双栏宽度
    \caption{Difference between long-input and long-output LLMs.}
    \label{fig:long-output_LLMs}
\end{figure}

\paragraph{The Case for Prioritizing Long Output.}
While the focus on long-context LLMs has primarily centered on processing extended input contexts, comparatively less attention has been given to long-output generation. This is surprising, given the growing number of applications requiring extended, coherent, and contextually rich text. Recent studies reveal significant performance limitations in existing models when tasked with generating long-form content beyond thousands of words~\cite{wu2024longgenbench,bai2024longwriter,ye2025longprocbenchmarkinglongcontextlanguage,tu2025longwriter}. 
This paper advocates for a shift in research priorities for foundational LLMs, urging researchers to focus on this relatively unexplored area of long text generation.  Several real-world applications, such as novel writing, long-term planning, and complex reasoning, require generating long texts exceeding 4,000 tokens (approximately 2,600 words) for successful task completion. Despite their importance, these applications have been significantly overlooked. These applications demand models capable of processing extensive contexts while producing high-quality, logically consistent outputs. We define these models, optimized for long-output tasks, as \model (see Figure \ref{fig:long-output_LLMs}).

%\paragraph{Prioritizing Long output in long-context LLMs.}
%The development of long-context language models (LLMs) has seen a surge in interest due to their wide range of applications and substantial commercial potential. However, the equally crucial aspect of long text generation has received comparatively less attention from both researchers and the industry (Subsection \ref{sec:2.2}). Existing discussions on context length predominantly emphasize a model's capacity to process long input contexts. However, these models often exhibit performance limitations during content generation, as evidenced by recent studies~\cite{wu2024longgenbench,bai2024longwriter,ye2025longprocbenchmarkinglongcontextlanguage}. \textbf{This paper advocates for a shift in research priorities for foundational models, urging researchers to focus on this relatively unexplored area of long text generation.} We contend that models for long text generation have been significantly overlooked, despite the prevalence of real-world applications requiring outputs exceeding 4,000 tokens (approximately 2,600 words) for effective task completion. Such tasks are not exceptional cases but rather represent a substantial portion of user needs where the generation of extended, coherent text is crucial. We define these models, capable of generating extended texts, as \model, as depicted in Figure \cc{\ref{fig:long-output LLMs}}.

\paragraph{Why Long-Output LLMs have been Overlooked?}
The limited progress in long-output generation can be attributed to three primary challenges. \textbf{\ding{172} Data limitations} pose a significant obstacle. Existing datasets for instruction-following tasks are predominantly composed of short input-output pairs, with only a limited number of high-quality datasets featuring long output sequences~\cite{bai2024longalign,xiong2024effective,chen2023longlora}. This scarcity of suitable data constrains both research and the practical application of long-output models. \textbf{\ding{173} Task execution complexities} add further difficulty. Generating long-form content, particularly for creative and structured tasks such as novel writing or article composition, requires models to maintain coherence and logical consistency across extended contexts. This level of complexity is significantly greater than what is required for shorter tasks~\cite{wu2024longgenbench,yang2024logu,tan-etal-2024-proxyqa}. \textbf{\ding{174} Computational cost constraints} present a substantial hurdle. The computational demand for generating long texts increases linearly in certain architectures~\cite{gu2023mamba,dao2022flashattention}. Furthermore, proprietary models often impose token limits (e.g., 4,096 or 8,192 tokens)~\cite{openai1,claude-3-5,googleai}, restricting their capacity to generate extended outputs. These combined challenges highlight the need for more targeted research and innovation to advance long-output LLM capabilities.

\paragraph{Why Care about the Long Output Domain?} Addressing the challenges of long-output LLMs is crucial for meeting real-world needs across various domains.  \textbf{\ding{172}}  Fields, such as healthcare, law, education, and media depend on long-form content for tasks such as generating research papers, drafting legal documents, and preparing detailed reports~\cite{zhao2024wildchat,chiang2024chatbot}. Long-output LLMs can bridge the gap in these areas by automating the production of coherent, high-quality content, thereby streamlining workflows. \textbf{\ding{173} Enhancing creativity and productivity} is another key benefit. Long-output LLMs facilitate the co-authoring of extensive works such as novels and academic papers, reducing the time and effort required for content creation. This allows professionals to allocate more attention to higher-level tasks like analysis and ideation~\cite{atmakuru2024cs4,chiang2024chatbot}. \textbf{\ding{174} Advancing complex reasoning} is a critical contribution of these models. By exploring larger output spaces and enhancing capabilities in summarization and inference, long-output LLMs enable deeper analysis and support intricate reasoning processes. Together, these advancements underscore the transformative potential of long-output LLMs in addressing real-world challenges.

In a nutshell, designing the first generation of a truly large, foundational \model could be an immensely rewarding and exciting opportunity for many researchers and workflow.

\paragraph{Paper Organization.} This paper begins by defining the concept of \model and highlighting their underrepresentation in current research (Section \ref{sec:long-content LLMs}). It then reviews the current state of research on long generation (Section \ref{sec:Current State}) and explores practical applications (Section \ref{sec:application}). The paper then discusses challenges and opportunities for advancing \model (Section \ref{sec:challenge}), followed by alternative views on \model and long-output generation (Section \ref{sec:Alternative Views}). Finally, the paper concludes by advocating for a strategic research shift towards long-output generation (Section \ref{sec:conclustion}).

%\paragraph{Overview.}

%This position paper begins by defining the concept of \model and highlighting how \model have been largely overlooked in current research (Section \ref{sec:long-content LLMs}). Next, we review existing research on long generation (Section \ref{sec:Current State}) and explore the applications of \model (Section \ref{sec:application}). We discuss the challenges and opportunities of transitioning towards \model (Section \ref{sec:challenge}). Finally, we return to the central argument of this paper, advocating for a shift in focus within the NLP community towards long generation (Section \ref{sec:conclustion}).

% 我们在这个章节首先将给出Long-content lLMs的 具体定义（一个long-content llm 需要具备哪些能力)。然后基于上述定义，讨论实际需求中这部分的占比，以及根据实际统计表明被显著忽视啦

\section{Long-Output LLMs}
\label{sec:long-content LLMs}
In this section, we first explore the prevalence of tasks involving these models in real-world applications, highlighting the significant neglect of related research, as revealed by statistical data. Next, we define \model and outline the requirements for a model to qualify as such

%In this section, we first provide a detailed definition of \model and outline the capabilities required for a model to be classified as such. Based on this definition, we then explore the prevalence of tasks that involve these models in real-world applications, highlighting the significant neglect of related research, as revealed by statistical data.

\begin{figure*}[t]
    \centering
    \begin{minipage}{0.45\textwidth}
        \centering
        \includegraphics[width=\textwidth, height=0.26\textheight, keepaspectratio]{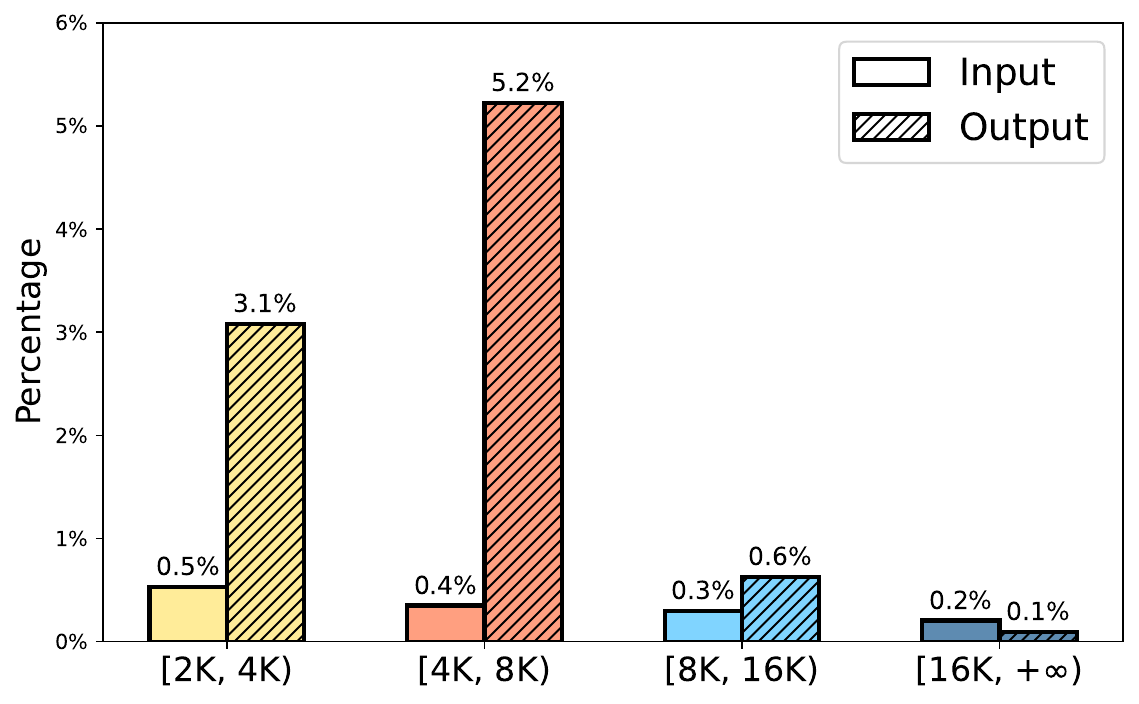}  
        \caption{Proportion of real-user demand: The aforementioned 2K (words) range refers to the interval [2K, 4K), and similarly for the other ranges. Solid color fill for input demand, slash fill for output.}
        \label{fig:High_Demand}
    \end{minipage}
    \hfill
    \begin{minipage}{0.48\textwidth}
        \centering
        \includegraphics[width=\textwidth, height=0.24\textheight, keepaspectratio]{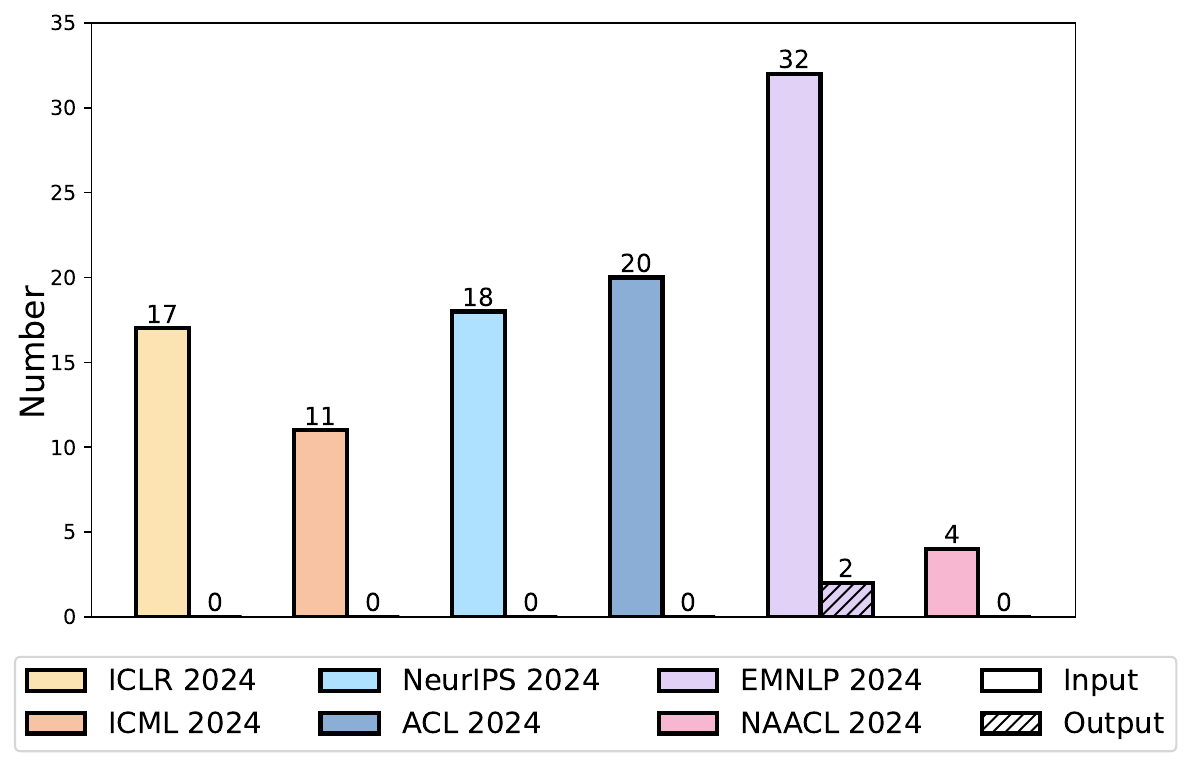}  
        \caption{ML and NLP Conf Long-context Research Trends Statistics (sorted by conference date). Solid color fill for Input-paper, slash fill for Output-paper.}
        \label{fig:Low_Focus}
    \end{minipage}
\end{figure*}

\subsection{High Demand, Low Research Focus}
\label{sec:2.2}

Despite the growing need for models capable of generating long-form content in real-world applications, significant gaps remain in targeted research. Tasks such as scientific writing, technical documentation, and AI-driven dialogues require models capable of producing coherent, high-quality outputs over extended spans. However, research has largely concentrated on input processing~\cite{an2024leval,hsieh2024ruler,vodrahalli2024michelangelo,reid2024gemini}, often neglecting the complexities of long-output generation~\cite{wu2024longgenbench,bai2024longwriter}. To substantiate this claim, we provide statistical evidence that underscores the limitations of prevailing research trends.

\paragraph{High Demand.} To quantify the increasing demand for long-output generation in natural language processing (NLP), we analyzed 100 K user requests from real-world scenarios, calculating the input-output length ratios using the Llama-3.3-70B model~\cite{dubey2024llama} and employed Few-shot learning predictions~\cite{brown2020language} to estimate the output length required for real-user queries\footnote{In Appendix \ref{app:query_stat}, describes the specific implementation of our statistics.}. Specifically, we examined four output lengths ranges—$[2K, 4K)$, $[4K, 8K)$, $[8K, 16K)$, and $[16K,+\infty )$ words—and compared them with the distribution of input lengths. The results reveal that demand for long-output generation exceeds equivalent-length inputs by more than 2-3 times in all cases except those with outputs greater than 16K
, with the ratio reaching nearly 15 times at $[4K, 8K)$ level(Figure~\ref{fig:High_Demand}). These findings underscore the necessity of long-output models for generating extensive, coherent content in practical applications\footnote{The user query statistics analysis of WildChat~\cite{zhao2024wildchat} is provided in Appendix \ref{app:wildchat}.}.

\paragraph{Low Research Focus:}
While significant progress has been made in research on large input processing, particularly for long-context models, the area of long-output generation remains underexplored. This imbalance is evident in our analysis of papers from leading ML (ICML, ICLR, NeurIPS) and NLP (ACL, EMNLP, NAACL) conferences in 2024\footnote{Appendix \ref{app:paperlist} includes all paper titles.}. Out of 104 papers addressing long-context tasks, only two specifically focused on long-output generation—a stark 102:2 ratio (Figure~\ref{fig:Low_Focus}). This imbalance focus is particularly concerning given the real-world demand for long-output models, which often surpasses the demand for long-input models.
%Although significant progress has been made in the research on large input processing, particularly in long-context models, long-output generation remains underexplored in the literature. This imbalance is concerning, especially in light of the growing demand for models capable of generating long, coherent content across a wide range of applications. To highlight this disparity, we analyzed the proportion of papers addressing long-input and long-output tasks at Top ML conferences (ICML, ICLR, NeurIPS) and NLP conferences (ACL, EMNLP, NAACL) in 2024\footnote{Appendix \ref{app:paperlist} show all paper title}. Our analysis reveals a significant research imbalance: over 98\% of papers focus on long-input tasks, while only 2\% are dedicated to long-output generation (102:2), as shown in Figure~\ref{fig:Low forcus}. This gap highlights the insufficient attention given to long-output models, despite their growing relevance in real-world applications. This is especially concerning given that the demand for long-output generation is at least as significant as, if not greater than, the demand for long-input processing.

\subsection{Defining Long-Output LLMs}

We propose that \model must satisfy two key requirements to effectively address the challenges of long-output generation:

\paragraph{Input: Context Handling Capabilities.} A model’s ability to handle extensive context is critical for producing coherent and contextually relevant outputs over long spans. As the length of generated text increases, the model must reference and integrate previous output segments to ensure logical flow and consistency. Transitioning from long-context models to \model requires enhanced capabilities in managing long-range dependencies and understanding complex, long-range contextual relationships. Benchmarks like LongBench-V2~\cite{bai2024longbench_2} demonstrate that these capabilities go beyond mere context processing, requiring a deep understanding of long contexts to answer questions accurately.

\paragraph{Output: Length and Quality of Generated Text.} While long-context models focus on processing extensive input (e.g., LLaMa 3.1~\cite{dubey2024llama} with 128K tokens or Gemini~\cite{reid2024gemini} with 1M tokens), \model prioritize the generation of long, coherent, and meaningful text. This involves producing outputs that span thousands or even millions of tokens while maintaining logical consistency, creativity, and relevance. Unlike traditional long-context models, which emphasize context size, \model excel at both the length and quality of the generated output, marking a significant step forward in natural language generation. This shift requires models to maintain coherence and quality across significantly longer and more complex content, setting \model apart as a foundational advancement.

In this paper, we establish a performance baseline, starting with 4K tokens (approximately 2.6K words)\footnote{The selection of 4K as the starting point for long-output LLMs is based on two key reasons. First, it aligns with the starting point established in long-context benchmarks such as Ruler~\cite{hsieh2024ruler}. Second, It aligns with the length of real-world requirements, as shown in Figure \ref{fig:High_Demand}.} as the effective length for long-content generation tasks.

\takeaway{Takeaway: We define \model as foundational LLMs specifically designed to excel at long-output tasks. While this definition allows for some flexibility, it broadly refers to large-scale language models capable of generating extended and coherent text.}

%\takeaway{Takeaway: We use the term \model\: to refer to a foundational LLMs designed for long-output tasks. While this definition is somewhat flexible, it generally refers to a large-scale language model that is capable of generating extended, coherent text outputs.}

\section{Current State of Long-Output LLMs}
\label{sec:Current State}
This section provides an overview of the current landscape of Long-Output LLMs, organized into three key areas: Data, Benchmarks, and Models. These dimensions collectively represent the core elements driving progress in the field—Data provides the foundation, Benchmarks set the evaluation standards, and Models showcase the cutting-edge advancements in long-output generation. 

%This section provides an overview of the current landscape of long-output LLMs, organized into three key areas: Data, Benchmarks, and Models. These dimensions are chosen because they collectively represent the core elements driving progress in the field—Data provides the foundation, Benchmarks set the evaluation standards, and Models showcase the cutting-edge advancements in long-output generation. 

% 对于pretrain阶段的数据，long-context LLMs 的 input 和 ouput 两个方面没有太大差异。而在SFT阶段存在加大的差异。
%  Xiong et al. (2023) proposes generating long instruction data by concatenating short instruction data and  Chen et al. (2023b) 和 longalign分别提供了  made their long instruction data, LongAlpaca-12k和 longalign。然而上述的long-context SFT数据大部分都是有着较长的input，而output较短。
% 最近Suri 通过Suri-I-ORPO  通过 backtranslated instruction 对于长文本生成指令，得到合适具有具有较长输出的的SFT的数据。而longwrite通过构建Agent，对于真实用户query，先生成plan，再分段进行生成得到合成的长output数据。XX 则是类似通过不断对于query的回答扩写得到答案。
%我们统计了上述的数据集的input output 平均长度在下表：XX 。\footnote{XX没有公开他们的训练数据}。

\subsection{Data}
\label{3.1_data}

During the long-context continual-pretraining phase, the datasets used for Long-input LLMs and Long-output LLMs overlap. However, significant divergence occurs during supervised fine-tuning. Early research primarily focused on datasets featuring extended input sequences while limiting output sequences to shorter lengths~\cite{xiong2024effective,xu2024chatqa}. For instance, datasets like LongAlpaca-12k~\cite{chen2023longlora} and LongAlign-10k~\cite{bai2024longalign} were designed for tasks such as summarization and question answering, where outputs length remained relatively constrained.

More recently, datasets have been developed to support the generation of longer, more detailed outputs. Suri~\cite{pham2024suri}, for example, employs backtranslation to transform long-content data into comprehensive instructions as input. Similarly, LongWriter-6k~\cite{bai2024longwriter} uses an agent-based methodology to generate a plan for user queries and then produces responses in segments, ensuring coherence in long-form outputs. Another work, Self-Lengthen~\citep{quan2024language}, uses iterative expansion to progressively extend responses through repeated elaboration, resulting in more detailed and lengthy outputs. 

This research represents a significant evolution in the field, as more datasets are specifically constructed to facilitate long-output generation—an essential capability for tasks that demand extensive reasoning or the production of extended text. Table \ref{tab:dataset_lengths} summarizes several key datasets' average input and output lengths, underscoring the growing focus on generating longer outputs to fine-tune long-context LLMs.

\begin{table}[t]
\centering
\begin{tabular}{lrr}
\toprule
\textbf{Dataset}  & \textbf{Input Length } & \textbf{Output Length} \\
\midrule

LongAlpaca-12k         & 5,945  & 218  \\
LongAlign-10k              & 12,134 & 169 \\
\hdashline

Suri            & 347  & 4,371  \\
LongWriter-6k              & 262  & 5,333 \\
\bottomrule
\end{tabular}
\caption{Comparison of Average Input and Output Lengths (words) for Long-Context SFT Datasets.}
\label{tab:dataset_lengths}
\end{table}

% 不同于关注于早期的long-context benchmark（关注于长input (>16K token)) ,也不同于一些long-form generation 专注于中等长度的output（关注于中等长度的output（~1K token）。 
% long-output llms的所需求的benchmark希望能评测更长的output的质量(>8K token(~5k word)
% 目前的对于如此之长output 用人工评价是困难的。现有的方法共有3类。
% 1，基于rule-base的，这种更多的是在要求model可以输出对应长度要求的文章，对于单词长度进行评估。评估方向比较单一。
% 2，基于LLMs-base的，这两种工作一种是使用LLMs直接对于整体进行评估，另一种会对于任务生成固定的checklist，来评价model是否完成要点。这两种方式都需要对output整体进行评估，对于LLMs具有较高要求。
% 3，基于分段的评估。LongGenBench要求LLMs进行结构化的输出，从而可以分段的评估long-output，使评估就有可解释性。但是只限于能结构化输出的任务，无法全面评价。
\subsection{Benchmarks and Evaluation}
Long-output benchmarks for long-context LLMs are designed to assess both the length and quality of outputs exceeding 4K tokens (approximately 2.6K words). These benchmarks differ significantly from traditional benchmarks that primarily focus on processing long input contexts ($\geq$16K tokens) \cite{needleinhaystack,hsieh2024ruler,bai2024longbench_2} or generating moderate-length outputs of around 1K tokens \cite{XuPSPFAC20,StelmakhLDC22,XuSW22,tan-etal-2024-proxyqa}. The unique challenge of long-output benchmarks lies in evaluating coherence, depth, and overall quality, where manual assessment becomes infeasible due to the extensive length of the generated text. To address this, three primary evaluation approaches have been developed.

The first approach is rule-based evaluation, which focuses on verifying output length by counting tokens or words \cite{bai2024longwriter,quan2024language,liu2024longgenbench}. While this method ensures compliance with predefined length requirements, it provides little to no insight into the quality, coherence, or depth of the generated content. Consequently, it is limited in its ability to offer a holistic evaluation of long-output models.

The second approach is LLM-based evaluation, which leverages the capabilities of LLMs to evaluate outputs in two distinct ways. One method involves using an LLM to assess the entire output holistically~\cite{bai2024longwriter,quan2024language,ye2025longprocbenchmarkinglongcontextlanguage}, while the other relies on a predefined checklist to determine whether the output meets specific criteria~\cite{pham2024suri,que2024hellobench}. Although LLM-based evaluations provide more comprehensive insights compared to rule-based methods, they are computationally expensive and heavily reliant on the model’s ability to understand and evaluate long, complex texts.

The third approach is segment-based evaluation, exemplified by frameworks like LongGenBench~\cite{wu2024longgenbench}. This method divides the output into smaller, more manageable segments, allowing for detailed and interpretable assessments of each portion. However, this approach is best suited to tasks that involve structured outputs and is less applicable to unstructured or narrative-based long-output tasks.

\subsection{Models}
While many recent models claim strong long-context capabilities, they often focus on handling long inputs rather than generating extended outputs. Benchmarks like LongGenBench~\cite{wu2024longgenbench} and LongWrite-Ruler~\cite{bai2024longwriter} reveal that current models struggle to maintain quality and coherence in outputs exceeding 4,000 tokens\footnote{
The details of the results from existing models are presented in Appendix \ref{app:poor_performance}.}. This limitation persists despite advancements in model architectures and training methods. 

%\roy{Do we want to have a figure or chart to illustrate this? It can be in appendix also. We can mentioned in a more technical terms what did it not do well.}

Three models demonstrate potential in generating extended outputs: \citet{bai2024longwriter}, \citet{pham2024suri}, and \citet{quan2024language}. These models share common methodologies, including the use of specialized datasets (as outlined in Section \ref{3.1_data}) and fine-tuning techniques to optimize long-output performance. Additionally, approaches like Direct Preference Optimization (DPO) \cite{rafailov2024direct} are used to refine output length control. However, despite these innovations, current models still face significant challenges in generating coherent, high-quality outputs at longer lengths, as evidenced by suboptimal performance on benchmarks like LongGenBench~\cite{wu2024longgenbench} and LongBench-Write~\cite{bai2024longwriter}.

%\takeaway{Takeaway: The pioneering efforts in long-output LLM research have laid the groundwork for this promising field. These early contributions highlight enormous potential and opportunities, positioning long-output generation as a critical area for further exploration and advancement.}

%Although many recent models assert strong long-context capabilities, these claims often focus on the models' ability to handle long inputs rather than generating extended outputs. Benchmarks like LongGenBench~\cite{wu2024longgenbench} and LongBench-Write~\cite{bai2024longwriter} highlight a significant issue: current models struggle to generate coherent, high-quality text when the length exceeds 4,000 tokens. This challenge persists despite advancements in model architectures and training methods.

%Three models demonstrate potential in generating extended outputs: \citet{bai2024longwriter}, \citet{pham2024suri}, and \citet{quan2024language}. These models follow a common approach: constructing specialized datasets as outlined in Subsection \ref{3.1_data}, followed by fine-tuning with long-context LLMs. Additionally, they use techniques like DPO~\cite{rafailov2024direct} to further refine output length control. Although these models can generate text of adjustable lengths, they fall short of achieving high performance on LongGenBench, struggling to maintain both quality and coherence in longer generations.

\takeaway{Takeaway: We greatly appreciate the early work on long-output LLMs. Their insightful discoveries have identified this direction as one with enormous potential and opportunities, making it a promising area for further research.}

%We greatly appreciate the early work on long-output LLMs. Their insightful discoveries have identified this direction as one with enormous potential and opportunities, making it a promising area for further research.

\section{Real-World Application}
\label{sec:application}
\subsection{Creative Writing Task}
% 由于长度的扩展，LLMs将不仅仅局限于原本较为短的长度create writing task。
% 可以尝试更加具有挑战性的应用场景，我们认为有下面几个领域能代表着，long-output的潜在应用场景
% 1，复杂且规范的写作任务，例如学术写作，法律文书。我们可以通过\model的实现而不局限于仅仅写个简单的email 或者 写会议paper一小段内容，而是直接输出整体。（这样的影响是什么，比如更深层次的解放人们的重复性工作。）
% 2，长篇的富有创意性的写作任务，例如写儿童读物，科幻小说。作家等文字工作者，可以通过\model直接获得整篇的创意，或者直接在此基础上进行refine。从而避免，多次分章节输出的所带的不连贯等问题。（这样的影响是什么，比如对于应对与更加广泛的创意工作。）
% 3，对的复杂的规划决策问题，例如设计领域，行程规划等。在这种情况下\model能具有更好的全局考虑，从而更出完整的规划方案，即使这种决策问题的输出token超过4K token。（这样的影响是什么，比如更好的且更加完整的方案。）
% 总结：总结来说，我们认为通过long-output llm能进一步的扩展现有LLMs的应用场景，尤其在create writing和planing 上。

The advancement of \model significantly broadens the scope of creative writing applications beyond traditional short-form tasks. This expansion facilitates the addressing of more complex and demanding writing scenarios, thereby demonstrating the versatile potential of \model in real-world contexts.

First, \model excel in generating complex and standardized documents, such as academic papers and legal documents. Unlike traditional models, which are often restricted to producing brief emails or small sections of a report, \model are capable of composing comprehensive, coherent documents in their entirety. This automation not only enhances efficiency by handling repetitive writing tasks but also frees up professionals to focus on higher-level analytical and decision-making activities, ultimately improving productivity and output quality.

Second, \model are particularly adept at facilitating creative writing endeavors, including genres such as children’s literature and science fiction. Writers can utilize \model to generate complete narratives, refine existing drafts, or maintain consistency and coherence across an entire work. This capability mitigates the common issue of disjointedness when generating content in smaller segments, allowing for a seamless creative process. By enabling large-scale, coherent content creation, \model support authors in realizing ambitious creative projects that would otherwise require significant time and effort.

In addition, \model contribute to complex planning and decision-making tasks, such as project design or itinerary creation. By generating detailed and holistic plans that consider multiple factors, these models ensure comprehensive and integrated solutions. This capability is particularly valuable for scenarios where the output exceeds typical token limits (e.g., 4K tokens), providing more thorough and contextually informed outcomes.

In summary, \model extend the functionality of existing language models by enabling the generation of extensive, coherent, and high-quality content across creative and strategic domains. This transformative potential positions \model as essential tools for automating and enhancing complex writing and planning tasks, driving innovation and efficiency across professional fields.

\subsection{Long Chain-of-Thought Task}
% \hzy{zhiqiang is all your need}

One of the most impactful applications of long-output generation is its ability to support long chain-of-thought (CoT) tasks, which require extended sequences of reasoning to solve complex problems. These tasks serve as a key benchmark for evaluating and advancing the capabilities of LLMs.

The long CoT approach, as exemplified by the OpenAI o1 model~\cite{openai2024openaio1card}, has demonstrated remarkable success across a range of domains. In mathematics, for instance, this method enables LLMs to tackle challenging problems, such as those encountered in Math Olympiads, while also excelling in tasks like complex code generation. These achievements underscore the transformative potential of long CoT techniques in domains that demand rigorous, systematic reasoning.

A critical enabler of long CoT success is the advancement of long-context scaling~\cite{kimi2025}. Complex reasoning tasks often result in extended outputs, requiring models to effectively manage both lengthy input and output sequences while maintaining coherence, relevance, and accuracy throughout. This necessitates innovations in scaling techniques to accommodate extended sequences without compromising performance. By refining these techniques, researchers ensure that LLMs can continue to perform effectively as the complexity and length of their outputs increase.

The progress in long CoT training highlights the importance of prioritizing long-output generation in LLM research. By tailoring models to meet the demands of long CoT applications, researchers unlock new possibilities in areas such as advanced problem-solving, strategic planning, and decision-making. As the field of long-output generation continues to evolve, its integration with long CoT tasks will play a pivotal role in shaping the future capabilities of LLMs, enabling them to address increasingly complex challenges across disciplines.

\section{Challenges and Opportunities}
\label{sec:challenge}
This section explores the challenges in advancing long-output large language models (\model) across three key areas: Data, Benchmarks, and Models. Additionally, it highlights opportunities to address these challenges and drive progress in the field.
%This section provides an overview of the challenge of shift  Long-output LLMs, organized into three key areas: Data, Benchmarks, and Models. 

\begin{figure*}[t]
    \centering
    \begin{minipage}{0.48\textwidth}
        \centering
        \includegraphics[width=\textwidth, height=0.26\textheight, keepaspectratio]{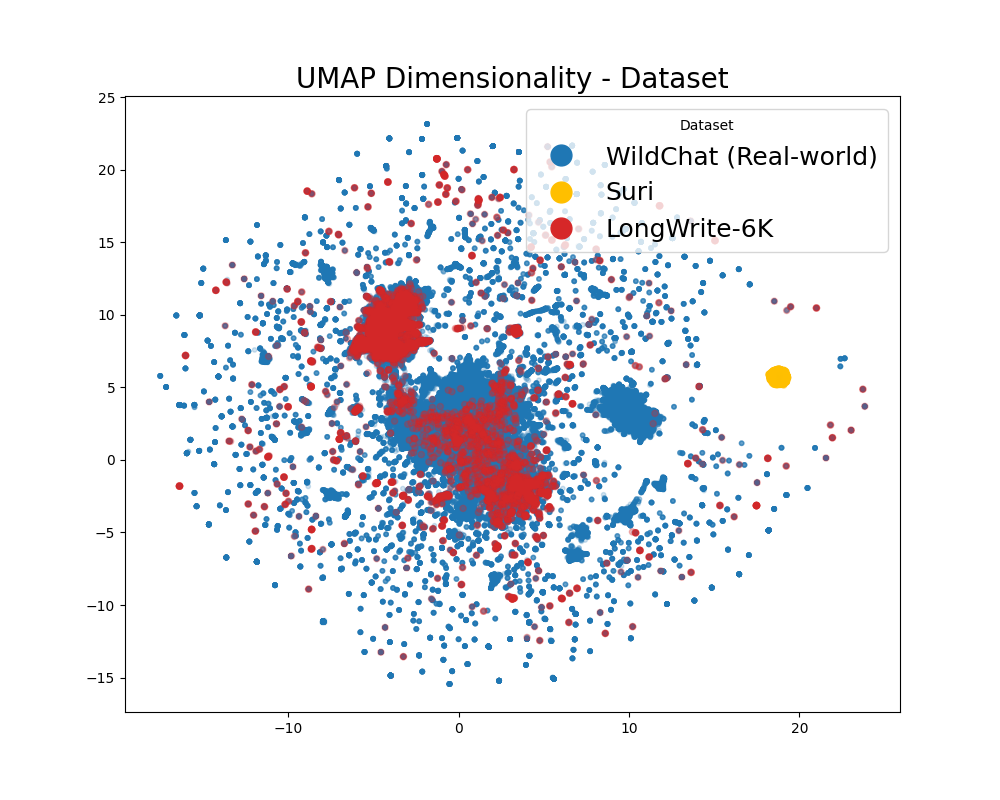}  
        \caption{UMAP visualization results for different SFT datasets. WildChat is derived from the long output demands of real users, filtered and referenced in Section \ref{sec:2.2}.}
        \label{Fig:dataset}
    \end{minipage}
    \hfill
    \begin{minipage}{0.48\textwidth}
        \centering
        \includegraphics[width=\textwidth, height=0.26\textheight, keepaspectratio]{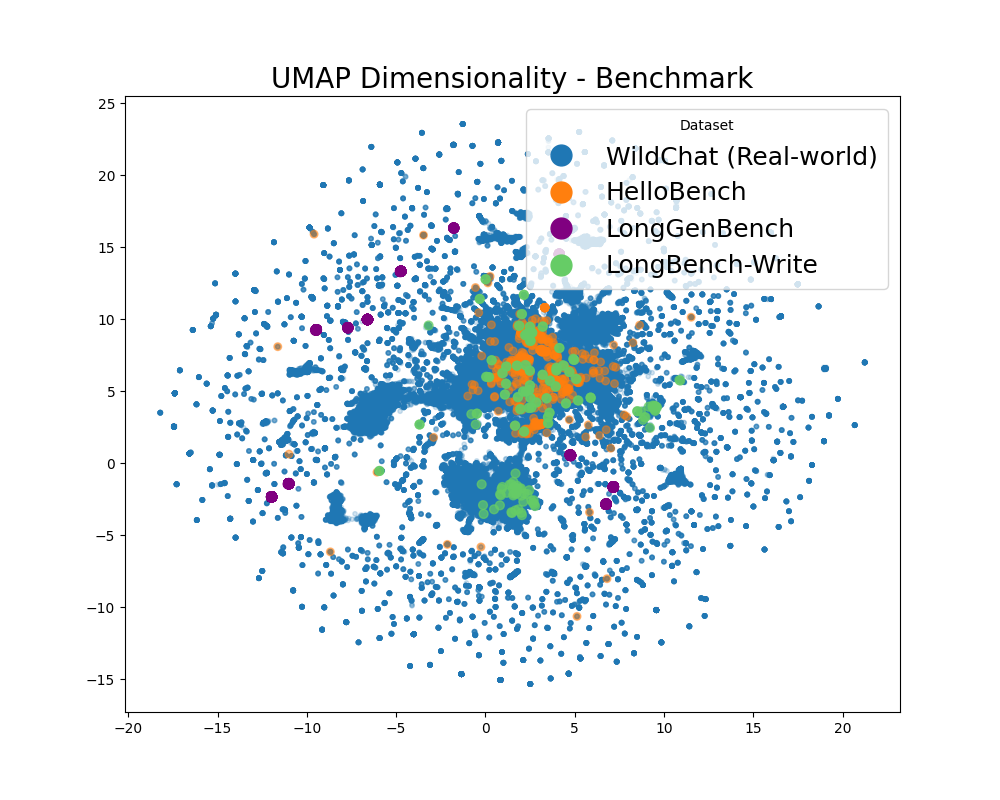}  
        \caption{UMAP visualization results for different benchmark. We use the instructions from the benchmark to evaluate whether the benchmark assesses a wide range of long-output demand.}
        \label{fig:benchmark}
    \end{minipage}

\end{figure*}

\subsection{Data}
As discussed in Section \ref{3.1_data}, current supervised fine-tuning (SFT) datasets for long-output tasks, such as LongWriter-6k~\cite{bai2024longwriter} and Suri-30K~\cite{pham2024suri}, face significant limitations. These challenges can be categorized into two main areas: user demand alignment and reliance on synthetic data.

% 根据section \ref{3.1}的分析，目前有两个开源的longwriting-6k和suri-30K的long-output SFT训练数据，但是他们还是有着一些limitation。在这里，我们将从长度分布 和 input需求匹配两个维度来说明现有的Long-output SFT数据无法满足需求。
% 用户需求判定（input）： 我们汇总了wild-chat的真实用户需求（作为真实场景的用户input，和suri和longwrite的input 进行对比，将其通过debert进行意图embedding，然后通过T-sne进行意图可视化。如图xx所示，Longwrite仅仅覆盖率极少的一部分用户需求，至于Suri所覆盖的范围更加偏少。因此现有数据构建是脱离现有模型训练的数据需求的。
% 长度分布：上述两个模型都宣称自己是一个面向长输出的llms，然而目前的两个数据集并没有包含所有所需求的数据分布，如图Fig X a）所示。因此现有的data无法满足所有用户需求，我们只能期待\model对于这种Out of distribution有着一定的泛化能力。
% 合成数据：由于天然长数据的稀缺性，有部分的工作使用合成的数据训练long-context LLM，例如longskywork，FILM等。然而经过我们的测试，经过合成数据的long-context LLMs并不适用于long-output任务,as show as tabel XXX。我们认为是由于现有的合成数据构建的上下文依赖关系，与真实场景相差极大，虽然在input部分的benchmark上表现出色，但是并不适用于long-output。

% 机会：现有的long-output SFT data有着诸多不足，因此如何开发高质量的数据是一个公开的问题和一个巨大的机会。我们任务，未来可以进一步的探索现有的通过Agent的方式以生成更高质量的数据方式以及构建更加大型的instruction tuning dataset\footnote{现有的SFT size 和 传统的SFT数据集存在较大的差距}。同时探究如何高效的进行数据合成而不影响long-output能力。

%As analyzed in Section \ref{3.1_data}, two publicly available supervised fine-tuning (SFT) datasets for long-output tasks: LongWriter-6k~\cite{bai2024longwriter} and Suri-30K~\cite{pham2024suri}. This section examines the limitations of these datasets from three key perspectives: user demand matching, length distribution, and the use of synthetic data.

\paragraph{User Demand Alignment:} Real-world user demands are often not reflected in the inputs provided by existing datasets. Our analysis of user Long-Output demands from WildChat~\cite{zhao2024wildchat}\footnote{Appendix \ref{app:query_stat} provides a detailed description of the specific implementation for obtaining long-output user requests.} compared these inputs with those in LongWriter and Suri, using all-mpnet-base-V2~\cite{reimers-2020-multilingual-sentence-bert} for embedding and T-SNE for visualization~\cite{van2008visualizing}. As shown in Figure \ref{Fig:dataset}, LongWriter only partially aligns with user demands, while Suri shows minimal overlap due to its synthetic instruction generation. This misalignment suggests that models trained on these datasets may struggle to generalize effectively to real-world scenarios, highlighting a critical gap in meeting user needs.

\paragraph{Synthetic Data:} Given the scarcity of natural long-text data, synthetic datasets such as Longskywork~\cite{zhao2024longskyworktrainingrecipeefficiently} and FILM~\cite{an2024makellmfullyutilize} have been widely used. However, as demonstrated by \citet{wu2024longgenbench}, synthetic data often introduces artificial dependencies that fail to capture the nuanced contextual relationships of real-world text. While these datasets may support performance on input-centric benchmarks, they fall short in enabling coherent and meaningful long-form text generation.

%\textbf{Synthetic Data:} Given the limited availability of natural long-text data, numerous studies have employed synthetic data to train long-context LLMs, such as Longskywork~\cite{zhao2024longskyworktrainingrecipeefficiently} and FILM~\cite{an2024makellmfullyutilize}, among others. However, \citet{wu2024longgenbench} demonstrate that LLMs trained on synthetic data are not well-suited for long-output tasks. This is due to the substantial differences in contextual dependencies between synthetic and real-world data. While these models may perform adequately on input-centric benchmarks, they have difficulty generating coherent long-form outputs. The artificial dependencies in synthetic data do not adequately capture the intricate and nuanced relationships present in natural language, leading to suboptimal performance in tasks that require coherent, long-form outputs.

\paragraph{Opportunities:} The limitations of current datasets create significant opportunities for innovation. First, real-world data collection, through collaborations with domain experts and industries, can yield high-quality, natural long-form datasets that better align training data with user demands. Second, agent-based approaches can simulate real-world scenarios, generating diverse, intent-rich data and larger instruction-tuning datasets\footnote{Current SFT datasets are significantly smaller than traditional ones.}. Third, hybrid approaches that combine synthetic and real-world data can balance scalability with contextual richness. Fourth, data augmentation techniques, such as iterative refinement and backtranslation, can enhance both dataset diversity and model robustness. Addressing these data-related challenges is critical for developing more robust and capable long-output LLMs.

\subsection{Benchmarks}

Long-output benchmarks aim to evaluate both the length and quality of outputs exceeding 4K tokens ($\approx$ 2.6K words). However, they face several challenges that hinder their effectiveness, particularly in model evaluation and applicability.

%Evaluating the performance of large language models (LLMs), particularly in long-output generation, presents significant challenges that remain largely unaddressed. These challenges fall into two main categories: the limited scope of existing benchmarks and the complexities associated with evaluating output quality.

\paragraph{Limited Scope of Benchmarks.} Similar to the gap between training data and real-world user demands, a significant mismatch exists at the level of benchmarks, with the disparity often being even more pronounced. As illustrated in Figure \ref{fig:benchmark} through UMAP results\footnote{The two  benchmarks have the same abbreviation, "longGenBench." In this study, we use \citet{wu2024longgenbench}, as the concatenation of GSM8K and MMLU in \citet{liu2024longgenbench} as long inputs differ significantly from the actual long-output demand.}, while benchmarks like LongWriter~\cite{bai2024longwriter} exhibit reasonable alignment with actual user needs, others, such as LongGenBench~\cite{wu2024longgenbench} and HelloBench~\cite{que2024hellobench}, demonstrate considerable divergence. This misalignment stems from the narrow scope of these benchmarks, which tend to focus on a limited subset of long-output tasks, leaving a vast range of real-world applications unaddressed. Consequently, the results derived from such benchmarks lack generalizability, impeding the development of models capable of tackling the diverse and complex requirements of long-output scenarios.

%Similar to the gap between training data and real-world user demands, a corresponding mismatch exists at the level of benchmarks, often more pronounced in many cases. As shown in Figure \ref{Fig:User Demand}, which illustrates UMAP results, while LongWriter\cite{bai2024longwriter} demonstrates reasonable alignment with actual user needs, other benchmarks, such as LongGenBench\cite{wu2024longgenbench} and HelloBench~\cite{que2024hellobench}, show significant divergence. This discrepancy arises from the fact that these benchmarks often focus on a narrow range of long-output tasks, failing to capture the full range of real-world applications. This narrow focus limits the generalizability of results based on these benchmarks, hindering the development of models capable of addressing a wider variety of long-output scenarios.

\paragraph{Evaluating Output Quality.} Assessing the quality of long-form text is challenging, as existing methods have significant limitations. Rule-based evaluations effectively measure specific aspects, such as mathematical reasoning~\cite{liu2024longgenbench} or instruction-following~\cite{wu2024longgenbench}, but fail to capture broader qualities like coherence, logical consistency, and narrative flow, providing only a partial picture of long-text generation quality. LLM-based evaluation methods offer broader capabilities but suffer from a lack of interpretability. They rarely explain why a text is rated poorly, making it difficult to identify areas for improvement. For instance, an experiment (Appendix \ref{app:mirror}) introduced a logical error into Snow White, where the shattered magic mirror continued to speak inconsistently\footnote{At a critical point in the story, a sentence is introduced that states the magic mirror has shattered and can no longer speak. As a result, any subsequent dialogue from the mirror becomes logically inconsistent.}. Most models failed to detect this flaw in the 3K-words version but succeeded with shorter text (300 words), highlighting current limitations in evaluating extended outputs. Additionally, LLM-based evaluations depend on the model’s ability to understand long texts, which remains an area of weakness~\cite{bai2024longbench_2}. High API costs further hinder their accessibility, limiting their adoption. These challenges emphasize the need for more interpretable, scalable, and cost-effective evaluation frameworks.

\paragraph{Opportunities.} 
The challenge of effectively evaluating \model remains an open issue. We propose that rule-based benchmarks can be enhanced by expanding their evaluation criteria to encompass qualities such as coherence, logical consistency, and creativity. While these qualities may lack readily available ground-truth labels, they can still be evaluated through the development of well-designed rules. For instance, creativity can be assessed using metrics like novelty and originality, in line with established methodologies \cite{zhao2024assessing}. To assess coherence, we propose potentially methods that dynamically construct a graph representing the narrative flow, allowing inconsistencies to be detected by identifying conflicts within the graph\footnote{For example, if a character is stated to have died, subsequent parts of the story should not depict them as living with their spouse.}. The current limitations of LLM-based methods primarily stem from their lack of interpretability and inherent difficulties in understanding long texts. We propose that, in developing LLMs for long-output generation, it would be beneficial to concurrently develop specialized reward LLMs tailored to specific tasks. This co-development strategy could enhance the accuracy, interpretability, and cost-effectiveness of evaluating long-output generation.

%The challenge of effectively evaluating \model remains an open issue. We propose that rule-based benchmarks can be enhanced by expanding their evaluation criteria to encompass qualities such as coherence, logical consistency, and creativity. While these qualities may lack readily available ground-truth labels, they can still be evaluated through the development of well-designed rules. For instance, creativity can be assessed using metrics like novelty and originality, in line with established methodologies \cite{zhao2024assessing}. To assess coherence, we propose exploring methods that dynamically construct a graph representing the narrative flow, allowing inconsistencies to be detected by identifying conflicts within the graph\footnote{For example, if a character is stated to have died, subsequent parts of the story should not depict them as living with their spouse.}. The current limitations of LLM-based methods primarily stem from their lack of interpretability and inherent difficulties in understanding long texts. We propose that, in developing LLMs for long-output generation, it would be beneficial to concurrently develop specialized reward LLMs tailored to specific tasks. This co-development strategy could enhance the accuracy, interpretability, and cost-effectiveness of evaluating long-output generation.

% 混合train问题
% inference overhead问题

\subsection{Train \& Inference}
% 混合train问题
% inference overhead问题

% 事实上，long-outout LLM的train及inference和long-context训练和inference差距不大，但是在我们的广泛尝试中，我们认为有如下两个挑战仍然并未解决。

% 混合数据train的 challenge：在训练过程中，不同侧重点的数据会相互影响彼此性能。例如在long-input训练的专有model似乎影响了其long-output的表现，反之亦然。与此同时，简单的混合训练数据并不能解决这个问题，反而会导致效果进行一部分降低。如图表X所示，在保证相同的数据量的时候，两种数据会在一定程度上相互制约，而无法具有合适的性能。

% inference time: 相比于long-input 的inference而言，由于long-output 的inference time会显著更高（在相同的length size下）。如表XX所示。long-output 的inference time几乎是long-input的X倍。这也是现有的所有api的output token 花费都几乎是input的4倍。同时由于之前没有更多针对于long-output的优化，可能会导致大部分的KV-cache 压缩的技术实效（如果这些加速方式都局限于input处理上。

% 机会：我们认为 对于混合训练的问题， 可以进一步通过数据配比，也可以尝试借鉴，Transfer learning 和 continal learning 的技术尝试解决。也可以对于两种数据的loss进行平衡。 对于inference time而言：应该对于long-output提出专有的优化方法，例如更合适的KV cache优化，也可以尝试其他的model架构，类似于mamba和kan。

The training and inference processes for long-output LLMs share similarities with those of long-context models. However, our extensive experimentation reveals two significant, largely unresolved challenges.

\paragraph{Model size:} A notable limitation of current long-output models is their reliance on smaller-scale architectures ($\leq$10B parameters) \cite{bai2024longwriter,pham2024suri}. While these models demonstrate the ability to handle long texts, scaling them to larger architectures capable of supporting more complex and higher-quality long-output generation remains a substantial challenge. Despite the advancements in state-of-the-art models, their performance highlights the pressing need for more sophisticated strategies to overcome scalability constraints. Larger model sizes, combined with efficient optimization techniques, are essential to fully realize the potential of generating extended, coherent, and high-quality text sequences.

%A significant limitation of these models is their reliance on smaller-scale LLMs ($\leq$ 10B). Although capable of handling long texts, scaling these models to support larger architectures remains a significant challenge. Current state-of-the-art models, while promising, underscore the need for more sophisticated strategies and larger model sizes to unlock the potential for generating long, high-quality text sequences.

% \paragraph{Mixed Data Training} One key challenge is training with mixed datasets, where data emphasizing different aspects can adversely impact the performance of each other. For instance, models specifically trained on long-input data tend to perform poorly on long-output tasks, and vice versa. Simply combining training data does not address this issue; in fact, it often results in a degradation of overall performance. As demonstrated in Table~\ref{tab:mixed_data_performance}, maintaining a consistent data volume reveals a trade-off between the two data types, preventing optimal performance in either case. This suggests that the model's ability to generalize is compromised, possibly due to conflicting learning signals or capacity limitations. The model may overfit to the dominant data type or fail to generalize across both. This phenomenon highlights the need for more advanced training strategies capable of integrating diverse data sources effectively.

\begin{figure}[t]
    \centering
    \includegraphics[width=0.85\columnwidth]{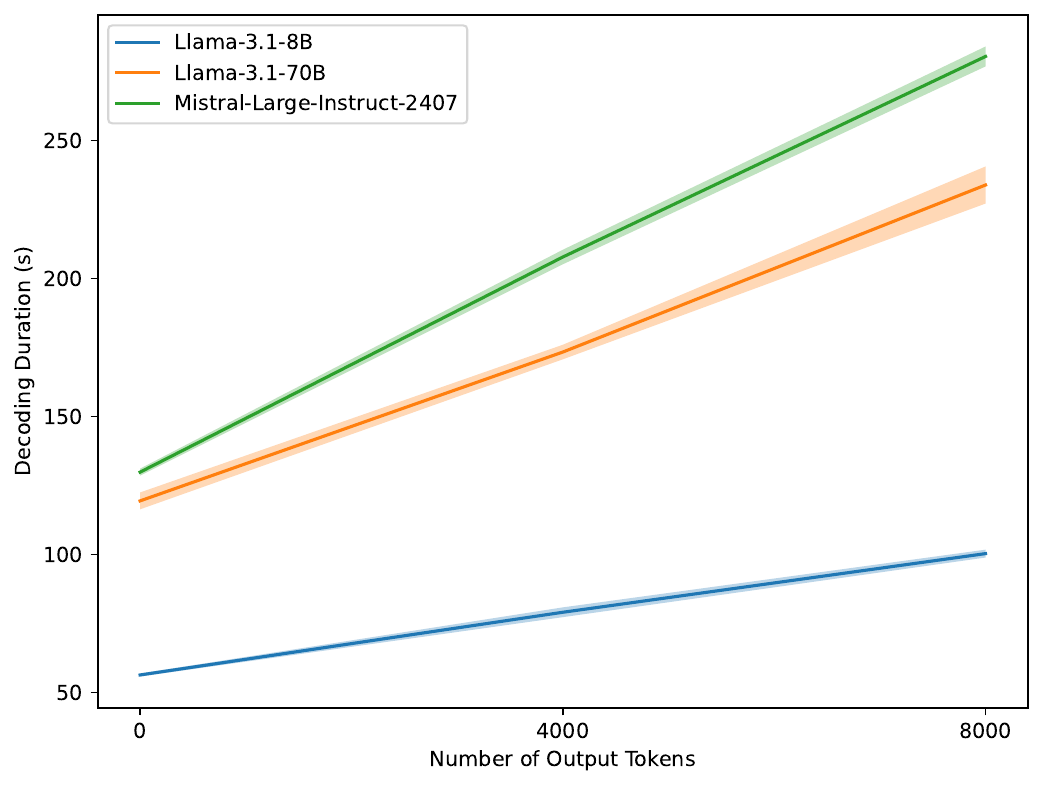}  % 图片宽度调整为栏宽的70%
    \caption{We set the total context length to 12,000 and gradually increased the proportion of output tokens.}
    \label{fig:Inference_result}
\end{figure}

\paragraph{Inference Time Overhead} Long-output inference incurs significantly higher time overheads than long-input inference, even for sequences of the same length. Long-output inference is often several times slower than long-input inference, as shown in Fig \ref{fig:Inference_result}. This observation aligns with the pricing models of existing APIs~\cite{googleai,openai1,glm2024chatglm}, where generating output tokens is typically more expensive than processing input tokens. The primary cause of this discrepancy lies in the iterative nature of output generation, where each token depends on the preceding ones. This sequential dependency limits parallelization and increases latency. Moreover, the lack of dedicated optimizations for long-output scenarios may reduce the effectiveness of existing KV-cache compression techniques, especially those focused on input processing. These factors collectively introduce a significant bottleneck in the efficient generation of long-output sequences, impeding the practical deployment of such models in real-time applications.

\paragraph{Opportunities} Several approaches can address the challenges in training and inference for long-output LLMs.  To mitigate inference time overhead, innovations in KV-cache management, parallelization techniques, and hybrid inference methods (e.g., combining autoregressive and non-autoregressive decoding) can significantly improve efficiency. Exploring architectures like Mamba~\cite{gu2024mambalineartimesequencemodeling}, LongMamba~\cite{anonymous2025longmamba}, and KAN~\cite{liu2024kankolmogorovarnoldnetworks} offers opportunities to optimize computational performance. Scaling beyond 10B parameters requires advancements in infrastructure, including distributed RL training frameworks~\cite{hu2024openrlhf} and low-memory optimization. Research into evaluating these solutions’ impacts on latency, coherence, and scalability metrics is essential for fully unlocking the potential of long-output models.

\section{Alternative Views}
\label{sec:Alternative Views}
While this paper advocates for prioritizing long-output generation in LLMs, there are alternative perspectives that challenge this position and propose different research priorities.

\paragraph{Long-Output Generation is Not Always Necessary:} An alternative view is that long-output generation might not be essential. Instead, we could use long-context LLMs for chained inference, where each output is used as input for the next step. This approach allows us to generate long texts step by step, avoiding the need for a single model to handle long outputs directly. By focusing on improving the coherence and efficiency of this chaining process, we can still achieve high-quality long outputs without the complexity of training models specifically for long-generation tasks. 

\paragraph{Long-Context Input Optimization Over Long-Output Generation:} A counterargument to the proposed focus on long-output generation is the continued prioritization of long-context input optimization. Critics may argue that the challenges associated with processing and understanding extensive input contexts are still unresolved, making them a prerequisite to achieving high-quality long outputs. Without robust mechanisms for efficiently handling long and diverse input, the coherence and logical consistency of extended outputs might remain unattainable. This perspective suggests that resources and research should remain directed toward input comprehension and scaling input capacity as the foundation for any downstream long-output tasks.

\paragraph{Computational Trade-offs:} The substantial computational costs associated with long-output generation—both during training and inference—might render this direction impractical for widespread use. Opponents may argue that research should focus on developing cost-effective techniques for moderate-length tasks, given that long-output LLMs might not be economically viable for most organizations and applications. These constraints could limit their adoption, especially in resource-constrained environments.

\paragraph{Evaluation Challenges as a Bottleneck:} Critics may emphasize that the lack of reliable evaluation metrics for long-output generation represents a fundamental bottleneck. Without robust evaluation frameworks, progress in this domain may remain speculative, with limited ability to measure real advancements or their relevance to user needs. This perspective supports prioritizing research in benchmark development before delving into long-output generation.

% \paragraph{Task-Specific Architectures Over Generalized LLMs:} An alternative approach advocates for task-specific architectures tailored for long-output generation in niche domains, such as legal document drafting or novel writing. Proponents argue that generalized LLMs may never achieve the specialized capabilities required for high-quality outputs across all contexts. Instead, research should focus on specialized systems optimized for particular tasks, ensuring higher performance and efficiency in real-world scenarios.

\section{Conclustion}
\label{sec:conclustion}
In this paper, we have defined and explored the potential of Long-Output LLMs. Despite their significant real-world applications, such models have yet to receive the attention they deserve in both academic research and practical implementations. As artificial intelligence and natural language processing continue to evolve, the ability to generate long-form content is becoming increasingly crucial, particularly for automated content creation, intelligent assistants, and complex information processing. We have identified the defining features, challenges, and emerging trends surrounding \model, calling for greater focus on advancing this domain. Future research should prioritize improving the quality, efficiency, and controllability of these models, while also developing new evaluation metrics and exploring diverse application scenarios. The development of \model has the potential to drive substantial change across industries, ushering in a new era of intelligent transformation.

\newpage

\nocite{}

\bibliography{example_paper}
\bibliographystyle{icml2025}

%%%%%%%%%%%%%%%%%%%%%%%%%%%%%%%%%%%%%%%%%%%%%%%%%%%%%%%%%%%%%%%%%%%%%%%%%%%%%%%
%%%%%%%%%%%%%%%%%%%%%%%%%%%%%%%%%%%%%%%%%%%%%%%%%%%%%%%%%%%%%%%%%%%%%%%%%%%%%%%
% APPENDIX
%%%%%%%%%%%%%%%%%%%%%%%%%%%%%%%%%%%%%%%%%%%%%%%%%%%%%%%%%%%%%%%%%%%%%%%%%%%%%%%
%%%%%%%%%%%%%%%%%%%%%%%%%%%%%%%%%%%%%%%%%%%%%%%%%%%%%%%%%%%%%%%%%%%%%%%%%%%%%%%
\newpage
\appendix
\onecolumn

\section{Proportion of real-user demand}
\label{app:query_stat}

\subsection{Input Length Statistics and Classification}

This subsection introduces a method for classifying the length of input text based on language and content size. The process involves two main steps: language detection and length calculation.

\begin{enumerate}
    \item \textbf{Language Detection}: Use regular expressions to identify the language of the text based on character counts. The text is classified as Chinese, Japanese, Korean, or English, depending on the dominant script.
    \item \textbf{Length Calculation}: The length of the text is calculated differently for different languages:
    \begin{itemize}
        \item For Chinese, Japanese, and Korean, character count is used.
        \item For English, word count is used.
    \end{itemize}
    \item \textbf{Length Classification}: Text is categorized into length buckets based on word count:
    \begin{itemize}
        \item 2K-4K: 2,000 to 4,000 words
        \item 4K-8K: 4,000 to 8,000 words
        \item 8K-16K: 8,000 to 16,000 words
        \item 16K+: More than 16,000 words
    \end{itemize}
\end{enumerate}

\subsection{Predicting User's Length Requirement with LLaMA 3.3-70B}

We describe a method for predicting the length of a user's input requirement using the LLaMA 3.3-70B model for few-shot learning. The process involves two main steps: predicting whether the input exceeds 2,000 words, and predicting the exact length requirement based on the first prediction.

\begin{enumerate}
    \item \textbf{Step 1: Predicting Length Exceedance (Prompt 1)}:
    The first prediction is made by checking whether the input exceeds 2,000 words. A carefully crafted prompt (Prompt 1) is provided to the model to predict if the content's expected word count will surpass the 2K threshold. The model utilizes few-shot learning with example inputs to classify the task into either ``above 2K'' or ``below 2K'' based on the nature of the input.
    \item \textbf{Step 2: Predicting Exact Length Requirement (Prompt 2)}:
    Once the model predicts whether the task exceeds 2,000 words, a second prediction is made to determine the exact length category. Based on the result from Step 1, Prompt 2 is designed to predict whether the content is in the 2K-4K, 4K-8K, 8K-16K, or 16K+ category. The model provides the final prediction by analyzing the contextual hints and the input length characteristics.
\end{enumerate}

\begin{tcolorbox}[size=title, opacityfill=0.1, title=\textbf{\textcolor{black}{Prompt-1}}]
\textbf{Guidelines:} \\
To determine whether the expected output will exceed 2000 words, consider the following factors:
\begin{enumerate}
    \item \textbf{Depth and Complexity:} Does the task require detailed explanations, in-depth analysis, or comprehensive coverage of complex topics?
    \item \textbf{Scope and Breadth:} Does the task cover multiple subtopics, perspectives, or extensive subject matter?
    \item \textbf{Structure and Sections:} Does the output need to include multiple sections such as introductions, literature reviews, methodologies, results, discussions, and conclusions?
    \item \textbf{Research and References:} Does the task require extensive research, citations, and referencing of multiple sources?
\end{enumerate}

\textbf{Response Format:}
\begin{itemize}
    \item Answer with either ``\#*\# Yes'' or ``\#*\# No''.
    \item Provide a concise justification based on the guidelines above.
\end{itemize}

\textbf{Example 1:} \\
Query: Is Sanskrit the oldest language? \\
Answer: This question requires a concise factual answer, not an extensive output. \#*\# No
\textbf{*** END}

\textbf{Example 2:} \\
Query: Create a detailed business plan for a new cat litter product. \\
Answer: Creating a detailed business plan involves multiple sections such as market research, product development, financial projections, marketing strategy, and competitive analysis, all of which require in-depth exploration and explanation. \#*\# Yes
\textbf{*** END}

.....

.....

Assess the following statement and decide whether the expected response is likely to require more than 2000 words. Answer with either ``\#*\# Yes'' or ``\#*\# No,'' and include a brief justification, like above example. \\
Query:  \cc{User Query} \\
Answer:
\end{tcolorbox}

\begin{tcolorbox}[size=title, opacityfill=0.1, title=\textbf{\textcolor{black}{Prompt-2}}]
\textbf{Guidelines:} \\
To estimate the expected length of the output, consider the following factors:
\begin{enumerate}
    \item \textbf{Depth and Complexity:} Does the task require detailed explanations, in-depth analysis, or complex reasoning?
    \item \textbf{Scope and Breadth:} Does the task cover multiple subtopics, perspectives, or an extensive subject matter?
    \item \textbf{Structure and Sections:} Does the output require multiple sections (e.g., introduction, literature review, methodologies, results, discussions, conclusions)?
    \item \textbf{Research and References:} Does the task require significant research, citations, or references to multiple sources?
    \item \textbf{Detail Level:} Is the task expected to be highly detailed, or can it be summarized concisely?
\end{enumerate}

\textbf{Response Format:} \\
- Choose the most likely word count category: ``Less than 2000 words'', ``2000 words'', ``4000 words'', ``8000 words'', or ``16000 words''. Using (\#\#\# Category: ``Chosen category'') as the response format. \\
- Provide a brief justification based on the guidelines above.

\textbf{Example 1: Less than 2000 words} \\
Query: Is Sanskrit the oldest language? \\
Answer: This is a factual question requiring a brief answer with no complex analysis or subtopics. Likely to be less than 2000 words. \#\#\# Category: Less than 2000 words \\
Explanation: Similar to a short blog post or brief news article, this task needs minimal detail and is concise. \textbf{*** END}

\textbf{Example 2: 2000 words (2000 to 4000 words)} \\
Query: Describe the key differences between classical and quantum computing. \\
Answer: This question requires moderate detail, comparing classical and quantum computing without exhaustive technical exploration. Likely to be around 2000 words. \#\#\# Category: 2000 words \\
Explanation: Similar to a moderate-length essay or a detailed blog post, this task covers key points with enough depth but remains manageable. \textbf{*** END}

.....

.....

\textbf{Example 5: More than 16000 words} \\
Query: Write a full-length book on the history of the Industrial Revolution, covering all major events, technological innovations, and global impacts. \\
Answer: This would require an in-depth exploration of the entire history of the Industrial Revolution, with detailed analysis across multiple chapters. Likely to be more than 16000 words. \#\#\# Category: 16000 words \\
Explanation: Similar to a book-length content, such as a thesis or encyclopedia entry, requiring substantial detail and coverage over multiple sections or chapters. \textbf{*** END}

Assess the following statement and decide what the expected output length is. Answer with the appropriate word count category and provide a brief justification. \\
Query: \cc{User Query}\\
Answer:
\end{tcolorbox}

\section{WildChat Long-Output Query Statistics and analysis}
\label{app:wildchat}
\begin{figure}[t]
    \centering
    \includegraphics[width=0.6\columnwidth]
    {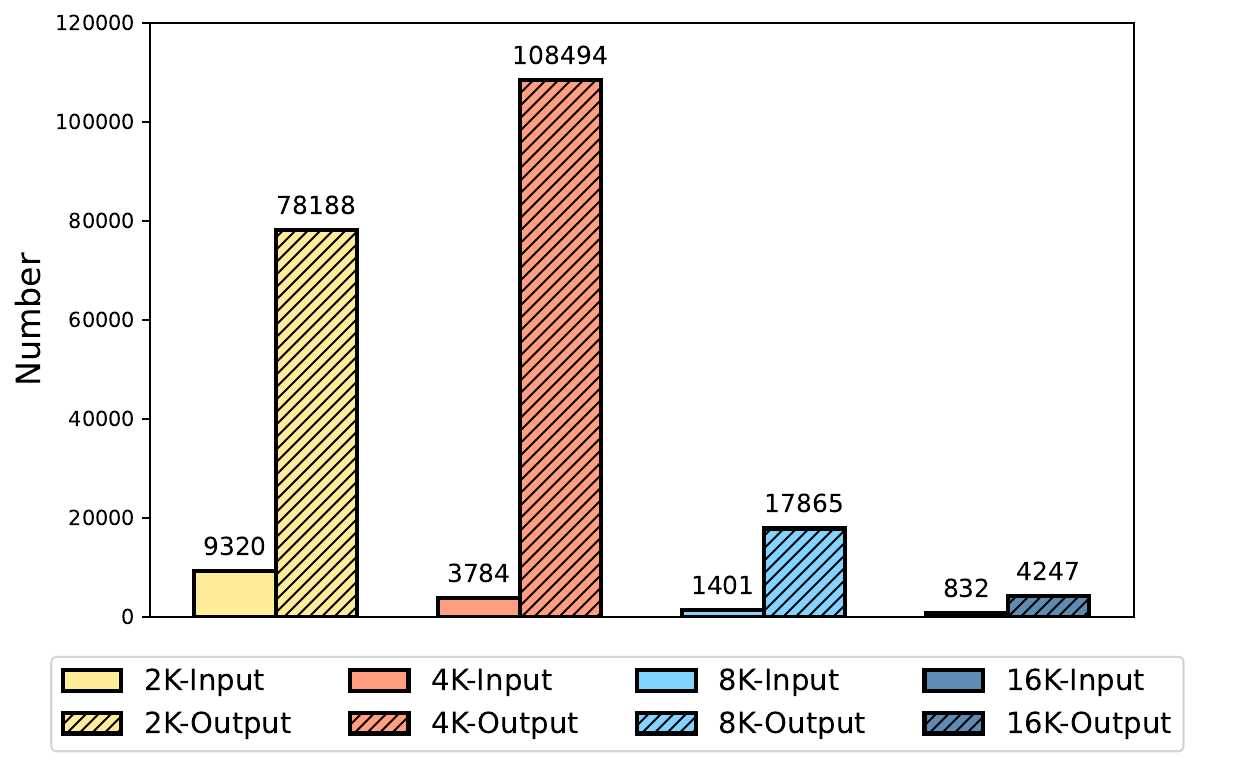}  % LaTeX默认的占位符，适应双栏宽度
    \caption{Proportion of real-user demand: The aforementioned 2K range refers to the interval [2K, 4K), and similarly for the other ranges. Solid color fill for input demand, slash fill for output demand in the Wildchat dataset.}
    \label{fig:wildchat_query_stat}
\end{figure}

We analyzed input-output length ratios using the WildChat: 1M ChatGPT Interaction Logs dataset~\cite{zhao2024wildchat} and the Llama-3.3-70B model~\cite{dubey2024llama}. Using few-shot learning~\cite{brown2020language}, we categorized output lengths into four ranges—$[2K, 4K)$, $[4K, 8K)$, $[8K, 16K)$, and $[16K,+\infty )$ words—and compared them to input lengths. The results show that demand for long-output generation is more than five times higher than for equivalent-length inputs, peaking at nearly 20 times at 4K-8K levels (Figure~\ref{fig:wildchat_query_stat}). However, due to the lack of file upload support in WildChat, the statistics for long-input queries are likely underestimated. 
\section{Long-context paper list}
\label{app:paperlist}
\subsection{ICML 2024}
\begin{enumerate}
    \item Linguistic Calibration of Long-Form Generations
    \item Short-Long Convolutions Help Hardware-Efficient Linear Attention to Focus on Long Sequences
    \item A Human-Inspired Reading Agent with Gist Memory of Very Long Contexts
    \item Memory Consolidation Enables Long-Context Video Understanding
    \item QUEST: Query-Aware Sparsity for Efficient Long-Context LLM Inference
    \item LLM Maybe LongLM: SelfExtend LLM Context Window Without Tuning
    \item LongRoPE: Extending LLM Context Window Beyond 2 Million Tokens
    \item Training-Free Long-Context Scaling of Large Language Models
    \item LoCoCo: Dropping In Convolutions for Long Context Compression
    \item Benchmarking and Building Long-Context Retrieval Models with LoCo and M2-BERT
    \item Data Engineering for Scaling Language Models to 128k Context
\end{enumerate}

\subsection{ICLR 2024}
\begin{enumerate}
    \item LongLoRA: Efficient Fine-tuning of Long-Context Large Language Models
    \item A Real-World WebAgent with Planning, Long Context Understanding, and Program Synthesis
    \item BooookScore: A Systematic Exploration of Book-Length Summarization in the Era of LLMs
    \item In-Context Pretraining: Language Modeling Beyond Document Boundaries
    \item Functional Interpolation for Relative Positions Improves Long Context Transformers
    \item RingAttention with Blockwise Transformers for Near-Infinite Context
    \item Hierarchical Context Merging: Better Long Context Understanding for Pre-trained LLMs
    \item FlashFFTConv: Efficient Convolutions for Long Sequences with Tensor Cores
    \item CLEX: Continuous Length Extrapolation for Large Language Models
    \item Retrieval Meets Long Context Large Language Models
    \item IceFormer: Accelerated Inference with Long-Sequence Transformers on CPUs
    \item PoSE: Efficient Context Window Extension of LLMs via Positional Skip-wise Training
    \item Efficient Streaming Language Models with Attention Sinks
    \item YaRN: Efficient Context Window Extension of Large Language Models
    \item Parallelizing Non-linear Sequential Models over the Sequence Length
    \item In-context Autoencoder for Context Compression in a Large Language Model
    \item HyperAttention: Long-context Attention in Near-Linear Time
\end{enumerate}
\subsection{NIPS 2024}
\begin{enumerate}
    \item Streaming Long Video Understanding with Large Language Models
    \item BABILong: Testing the Limits of LLMs with Long Context Reasoning-in-a-Haystack
    \item Perceiving Longer Sequences with Bi-Directional Cross-Attention Transformers
    \item Video Token Merging for Long Video Understanding
    \item Chain of Agents: Large Language Models Collaborating on Long-Context Tasks
    \item LoTLIP: Improving Language-Image Pre-training for Long Text Understanding
    \item Stress-Testing Long-Context Language Models with Lifelong ICL and Task Haystack
    \item MInference 1.0: Accelerating Pre-filling for Long-Context LLMs via Dynamic Sparse Attention
    \item MMLONGBENCH-DOC: Benchmarking Long-context Document Understanding with Visualizations
    \item An Efficient Recipe for Long Context Extension via Middle-Focused Positional Encoding
    \item Found in the Middle: How Language Models Use Long Contexts Better via Plug-and-Play Positional Encoding
    \item Mini-Sequence Transformers: Optimizing Intermediate Memory for Long Sequences Training
    \item Mixture of In-Context Experts Enhance LLMs' Long Context Awareness
    \item MMBench-Video: A Long-Form Multi-Shot Benchmark for Holistic Video Understanding
    \item StreamingDialogue: Prolonged Dialogue Learning via Long Context Compression with Minimal Losses
    \item InfLLM: Training-Free Long-Context Extrapolation for LLMs with an Efficient Context Memory
    \item LongVideoBench: A Benchmark for Long-context Interleaved Video-Language Understanding
    \item Rethinking Transformer for Long Contextual Histopathology Whole Slide Image Analysis
\end{enumerate}

\subsection{ACL 2024}
\begin{enumerate}
    \item L-Eval: Instituting Standardized Evaluation for Long Context Language Models
    \item Analyzing Temporal Complex Events with Large Language Models? A Benchmark Towards Temporal, Long Context Understanding
    \item LongLLMLingua: Accelerating and Enhancing LLMs in Long Context Scenarios via Prompt Compression
    \item Making Long-Context Language Models Better Multi-Hop Reasoners
    \item LongBench: A Bilingual, Multitask Benchmark for Long Context Understanding
    \item Landmark Embedding: A Chunking-Free Embedding Method For Retrieval Augmented Long-Context Large Language Models
    \item CoCA: Fusing Position Embedding with Collinear Constrained Attention in Transformers for Long Context Window Extending
    \item NextLevelBERT: Masked Language Modeling with Higher-Level Representations for Long Documents
    \item RelayAttention for Efficient Large Language Model Serving with Long System Prompts
    \item Marathon: A Race Through the Realm of Long Context with Large Language Models
    \item $\infty$Bench: Extending Long Context Evaluation Beyond 100K Tokens
    \item FinTextQA: A Dataset for Long-form Financial Question Answering
    \item Long Context is Not Long at All: A Prospector of Long-Dependency Data for Large Language Models
    \item M4LE: A Multi-Ability Multi-Range Multi-Task Multi-Domain Long-Context Evaluation Benchmark for Large Language Models
    \item Chunk, Align, Select: A Simple Long-sequence Processing Method for Transformers
    \item LooGLE: Can Long-Context Language Models Understand Long Contexts?
    \item Never Lost in the Middle: Mastering Long-Context Question Answering with Position-Agnostic Decompositional Training
    \item DocFinQA: A Long-Context Financial Reasoning Dataset
    \item SumSurvey: An Abstractive Dataset of Scientific Survey Papers for Long Document Summarization
    \item Found in the Middle: Calibrating Positional Attention Bias Improves Long Context Utilization
\end{enumerate}

\subsection{EMNLP 2024}
\begin{enumerate}
    \item LongEmbed: Extending Embedding Models for Long Context Retrieval
    \item Forgetting Curve: A Reliable Method for Evaluating Memorization Capability for Long-Context Models
    \item LUQ: Long-text Uncertainty Quantification for LLMs
    \item Leave No Document Behind: Benchmarking Long-Context LLMs with Extended Multi-Doc QA
    \item CItruS: Chunked Instruction-aware State Eviction for Long Sequence Modeling
    \item Attribute or Abstain: Large Language Models as Long Document Assistants
    \item Summary of a Haystack: A Challenge to Long-Context LLMs and RAG Systems
    \item AnaloBench: Benchmarking the Identification of Abstract and Long-context Analogies
    \item Where am I? Large Language Models Wandering between Semantics and Structures in Long Contexts
    \item FinDVer: Explainable Claim Verification over Long and Hybrid-content Financial Documents
    \item LONGAGENT: Achieving Question Answering for 128k-Token-Long Documents through Multi-Agent Collaboration
    \item Is It Really Long Context if All You Need Is Retrieval? Towards Genuinely Difficult Long Context NLP
    \item SEGMENT+: Long Text Processing with Short-Context Language Models
    \item One Thousand and One Pairs: A “Novel” Challenge for Long-Context Language Models
    \item LLoCO: Learning Long Contexts Offline
    \item Enhancing Post-Hoc Attributions in Long Document Comprehension via Coarse Grained Answer Decomposition
    \item Memorize Step by Step: Efficient Long-Context Prefilling with Incremental Memory and Decremental Chunk
    \item LongRAG: A Dual-perspective Retrieval-Augmented Generation Paradigm for Long-Context Question Answering
    \item Losing Visual Needles in Image Haystacks: Vision Language Models are Easily Distracted in Short and Long Contexts
    \item LongWanjuan: Towards Systematic Measurement for Long Text Quality
    \item LongHeads: Multi-Head Attention is Secretly a Long Context Processor
    \item Insights into LLM Long-Context Failures: When Transformers Know but Don’t Tell
    \item LongGenBench: Long-context Generation Benchmark
    \item Can’t Remember Details in Long Documents? You Need Some R\&R
    \item GraphReader: Building Graph-based Agent to Enhance Long-Context Abilities of Large Language Models
    \item More Bang for Your Context: Virtual Documents for Question Answering over Long Documents
    \item LongAlign: A Recipe for Long Context Alignment of Large Language Models
    \item Long Sequence Modeling with Attention Tensorization: From Memory-Efficient Design to Long-context OpenQA
    \item LSM1K: Large Scale Memory-based Dataset for Long Text Modeling
\end{enumerate}

\subsection{NAACL 2024}
\begin{enumerate}
    \item RST-LoRA: A Discourse-Aware Low-Rank Adaptation for Long Document Abstractive Summarization
    \item Ada-LEval: Evaluating long-context LLMs with length-adaptable benchmarks
    \item Effective Long-Context Scaling of Foundation Models
    \item WESOME: GPU Memory-constrained Long Document Summarization using Memory Mechanism and Global Salient Content
\end{enumerate}

\subsection{Long-output}
\begin{enumerate}
    \item Suri: Multi-constraint Instruction Following in Long-form Text Generation
    \item LongGenBench: Long-context Generation Benchmark
\end{enumerate}
\section{Broken Mirror Test}
\label{app:mirror}
We conducted a simple test, and almost all models~\cite{GPT-4o,dubey2024llama,GPT-4o-mini,reid2024gemini,glm2024chatglm} failed to correctly answer the Prompt-long (only O1~\cite{o1-preview} was able to answer correctly, but not stable). We identified logical errors in handling long text cases, while the models were almost always able to correctly respond to the Prompt-short.

\begin{tcolorbox}[size=title, opacityfill=0.1, title=\textbf{\textcolor{black}{Prompt-long (3089 words)}}]
\textbf{This is a fairy tale. Please carefully consider whether there are any logical issues in the following story. If there are, point out just one significant logical flaw?}

Once upon a time in midwinter, when the snowflakes were falling like feathers from heaven, a queen sat sewing at her window, which had a frame of black ebony wood. As she sewed she looked up at the snow and pricked her finger with her needle. 

.....

.....

Mirror, mirror, on the wall, Who in this land is fairest of all? To this the mirror answered: You, my queen, are fairest of all.

.....

.....

\cc{With that, she slammed the mirror with all her strength, shattering it into pieces, her fury burning in her heart. The mirror was instantly destroyed, unable to speak again.}

.....

.....

Back at home she asked her mirror: Mirror, mirror, on the wall, Who in this land is fairest of all? \red{It finally answered: You, my queen, are fairest of all.} \ding{55} 

.....

.....

\end{tcolorbox}

\begin{tcolorbox}[size=title, opacityfill=0.1, title=\textbf{\textcolor{black}{Prompt-short (390 words)}}]
\textbf{This is a fairy tale. Please carefully consider whether there are any logical issues in the following story. If there are, point out just one significant logical flaw?}

Once upon a time in midwinter, when the snowflakes were falling like feathers from heaven, a queen sat sewing at her window, which had a frame of black ebony wood. As she sewed she looked up at the snow and pricked her finger with her needle. 

.....

Mirror, mirror, on the wall, Who in this land is fairest of all? To this the mirror answered: You, my queen, are fairest of all.

When the queen heard the mirror say this, she shook and trembled with anger, shouting, "Snow-White must die." 

\cc{With that, she slammed the mirror with all her strength, shattering it into pieces, her fury burning in her heart.  The mirror was instantly destroyed, unable to speak again.}

"Tell me, mirror," she said, her voice softer now, "will I always be the fairest in the land?"

\red{"True beauty comes not from the surface, but from the soul," the mirror answered. "What you seek is fleeting. Find peace, and you will see a beauty that lasts beyond your own."} \ding{55}

The Queen’s eyes hardened again. "I do not seek peace. I seek power."

And so, with a cold smile, she turned away from the mirror, her heart set on a path that would lead her further from the light.

\end{tcolorbox}

\section{Current model in Long-output Benchmark}
\label{app:poor_performance}
\begin{figure}[t]
    \centering
    \includegraphics[width=0.6\columnwidth]{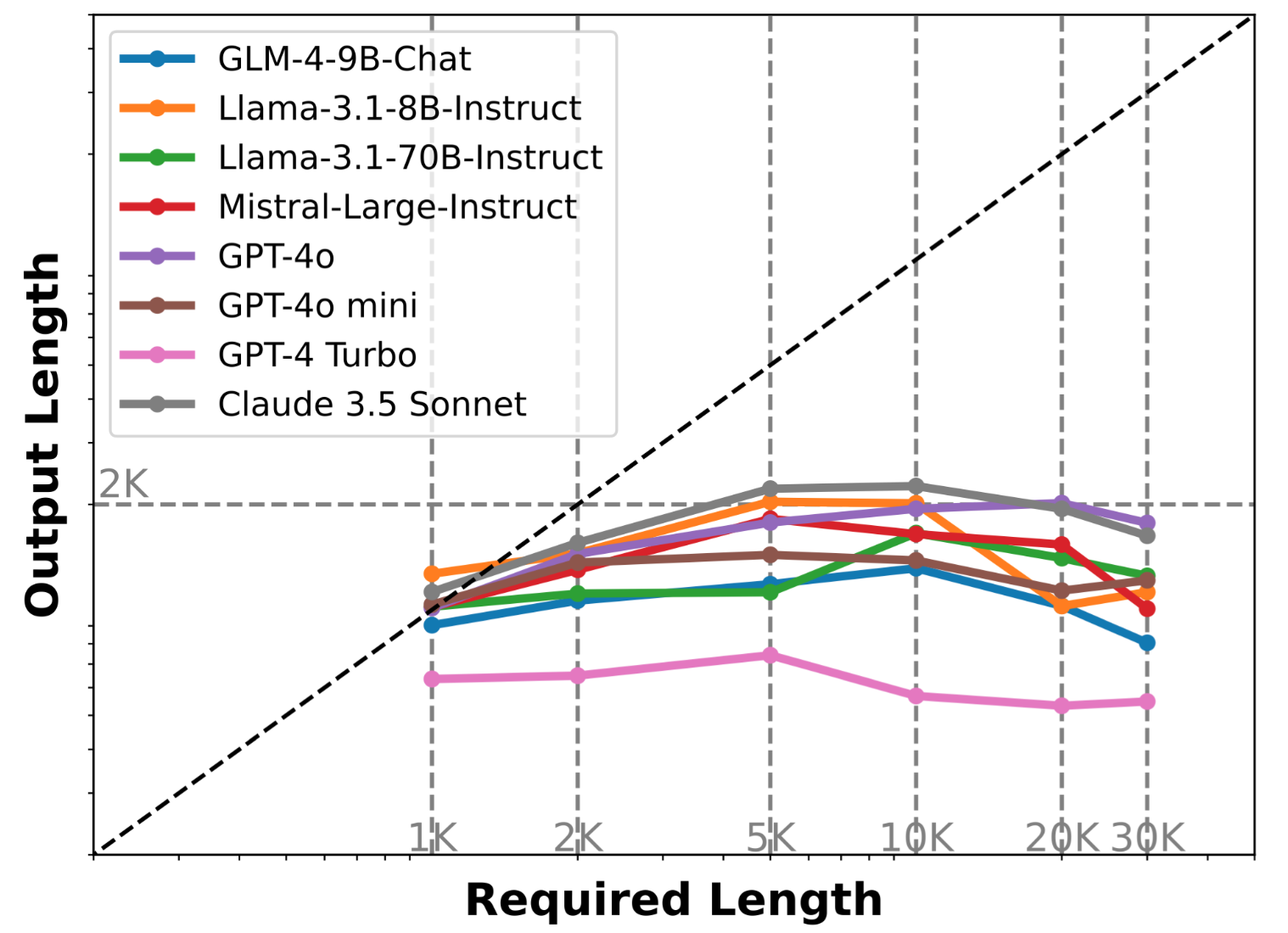}  % 80% 的宽度
    \caption{LongWriter-Ruler test demonstrates a
maximum output length limitation of approximately 2k words for all models tested.}
    \label{fig:poor_performance}
\end{figure}

The Fig \ref{fig:poor_performance} is from Longwrite~\cite{bai2024longwriter}, and it reveals that most existing LLMs are unable to meet the output requirements for long instructions.

% You can have as much text here as you want. The main body must be at most $8$ pages long.
% For the final version, one more page can be added.
% If you want, you can use an appendix like this one.  

% The $\mathtt{\backslash onecolumn}$ command above can be kept in place if you prefer a one-column appendix, or can be removed if you prefer a two-column appendix.  Apart from this possible change, the style (font size, spacing, margins, page numbering, etc.) should be kept the same as the main body.
%%%%%%%%%%%%%%%%%%%%%%%%%%%%%%%%%%%%%%%%%%%%%%%%%%%%%%%%%%%%%%%%%%%%%%%%%%%%%%%
%%%%%%%%%%%%%%%%%%%%%%%%%%%%%%%%%%%%%%%%%%%%%%%%%%%%%%%%%%%%%%%%%%%%%%%%%%%%%%%

\end{document}